\documentclass[letterpaper, 10 pt, conference]{ieeeconf}  

\usepackage{amsmath,amsfonts}
\usepackage{algorithmic}
\usepackage{algorithm}
\usepackage{array}
\usepackage[caption=false,font=normalsize,labelfont=sf,textfont=sf]{subfig}
\usepackage{bm}
\usepackage{textcomp}
\usepackage{stfloats}
\usepackage{booktabs}
\usepackage{url}
\usepackage{verbatim}
\usepackage{multirow}
\usepackage{graphicx}
\usepackage{cite}
\usepackage{siunitx}
\usepackage{xcolor}
\usepackage{hyperref}
\usepackage{colortbl}
\let\labelindent\relax 
\usepackage[inline]{enumitem}
\newcommand{\myparagraph}[1]{\par\noindent\textbf{#1.}}
\usepackage{eso-pic}

\newcommand\submittednotice{%
  \AddToShipoutPictureBG*{%
    \AtPageLowerLeft{%
      \parbox[b][1.5cm][c]{\paperwidth}{\centering\footnotesize
        This work has been submitted to the IEEE for possible publication. Copyright may be transferred without notice, after which this version may no longer be accessible.}%
    }%
  }%
}

\IEEEoverridecommandlockouts                              

\definecolor{customgray}{gray}{0.7} 

\title{\LARGE \bf
MobileNeTTT: a Lightweight DNN-based Time-To-Transit Estimator for Visual Navigation
}

\author{Chiara Boretti, Philippe Bich,
Manuel Morteo$^{\ddagger}$,
Roberto Tron$^{\ddagger}$, \textit{Senior Member, IEEE},
John Baillieul$^{\ddagger}$, \textit{Fellow, IEEE}
\thanks{$^{\ddagger}$M.M.,  R.T.\ and J.B.\ are with the College of Engineering at Boston University.
Support from various sources including the U.S.\ Office of Naval Research grants N00014-10-1-0952, N00014-17-1-2075, and N00014-19-1-2571 is gratefully acknowledged.}}

\begin{document}

\submittednotice
\maketitle
\thispagestyle{empty}
\pagestyle{empty}

\begin{abstract}

Time-to-Transit (TTT) is a perceptual cue, proxy for depth, that can be computed in the visual cortex, allowing animals to estimate the time remaining to move past features in the environment. We show how TTT can be estimated and used for bio-inspired navigation of a mobile robot equipped with a monocular camera.  The paper describes a lightweight neural network estimation method, named MobileNeTTT, that
(1) uses the full field of view and provides a rich array of visual feature TTT estimates;
(2) is relatively insensitive to camera rotation; and
(3) automatically adjusts to the geometry of the path and the presence of obstacles.
Because TTT is a coarse visual cue, it is far simpler to use for robot navigation than current large visual depth foundation models.  Our navigation model uses a 1.95M-parameter network that is two to three orders of magnitude smaller than well-known large vision foundation models, making it able to run in real time on edge hardware.  Simulation and actual laboratory experiments validate the proposed approach in dynamic and spatially complex environments. Code, data, and video are available.\footnote{\url{https://github.com/johnbaillieul/MobileNeTTT_VisionNavigation}}

\end{abstract}

\section{INTRODUCTION}

Studies of flying animals have revealed much about the biological sensory-motor systems that enable agile flight through complex, cluttered, and dynamic environments~\cite{srinivasanBiologyEngineeringInsect2012}. Vision is chief among these modalities, and a growing body of bio-inspired robotics has used optical flow to guide steering~\cite{decroonAccommodatingUnobservabilityControl2022}, building on the work of James J.~Gibson~\cite{gibsonPerceptionVisualWorld1950}.
We focus in particular on a visual cue that was originally called \emph{time-to-contact} (TTC), which has been shown to be a possible means by which a diving sea bird could estimate when it would enter the water \cite{leePlummetingGannetsParadigm1981,leeTheoryVisualControl1976}.
Since then, it has been a significant focus of research in perceptual psychology~\cite{fajenSteeringGoalEqualizing2001, vagnoniActionAbilityModulates2017}, and has been of specific interest for navigation~\cite{jiangInterculturalAnalysesTimetoCollision2015a, liEventAidedTimetoCollisionEstimation2024a}.
In~\cite{baillieulPerceptualControlLarge2019}, a concept closely related to TTC called \emph{Time-To-Transit} (TTT), or $\tau$, was proposed.  Leveraging this visual cue, it was shown that steering laws based on monocular camera perceptions of $\tau$ could reliably steer a mobile vehicle to mimic the navigation strategies observed in birds, bats, and bees~\cite{zhaodankongOpticalFlowSensing2013}.
\begin{figure}[]
    \centering
    \includegraphics[width=2.6in]{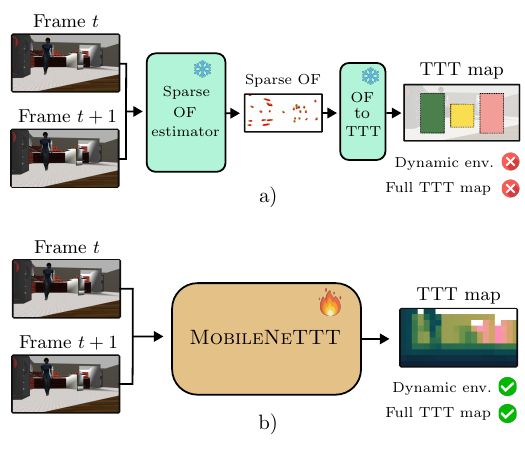}
    \vspace{-.4cm}
    \caption{In a) TTT estimation from sequential monocular images using sparse Optical Flow (OF), while in b) MobileNeTTT, which extends coarse TTT estimation across the entire image, enhancing estimation, maintaining real-time performance without imposing any motion constraint. }
\label{fig:mnett_base}
\end{figure}

To replicate the visual proficiency of animals in engineered systems, accurate TTT estimation is crucial. In~\cite{borettiVisualNavigationUsing2022}, authors used estimates based on sparse Optical Flow (OF) provided by the well-known Lucas-Kanade (LK) method (Fig.~\ref{fig:mnett_base}a).
Despite its success in moderately complex environments, the approach had several limitations:
\begin{enumerate*}
\item it provided an overly coarse TTT estimate (left, center, and right regions)
\item it imposed motion constraints to ensure reliable predictions, and
\item it failed to adapt the steering control laws to account for the geometry of the environment.
\end{enumerate*}

In this work, we propose MobileNeTTT (Fig.~\ref{fig:mnett_base}b), a lightweight neural-network-based method that addresses these limitations as follows:
\begin{enumerate}
    \item We systematically evaluate different architecture options,  comparing and improving on them to obtain a model that is well-tailored for TTT estimation.
    \item It estimates TTT over a broader and denser grid spanning the entire image. Thus, it ensures full visual field coverage while preserving the bio-inspired principle of coarse estimation.
    \item It does not rely on sparse optical flow. Instead, it leverages a DNN-based depth estimator that is independent of robot speed. The coarse estimation scale enables real-time performance: because TTT is a coarse cue, the per-pixel fidelity, sharp object boundaries, and occlusion handling that large depth and navigation foundation models are designed to deliver are unnecessary, and even counterproductive, for this task. This is precisely why a tiny network matches or surpasses far larger ones on the coarse estimation task (Sec.~\ref{sec:training}).
    \item It 
    automatically adapts control laws to the characteristics of the environment (e.g., curved path segments and corners and \emph{a priori} unknown locations of obstacles).
\end{enumerate}


\section{RELATED WORK AND GEOMETRY OF TTT}
\label{sec:related_works}
\myparagraph{Optical flow and visual cues for guiding movement}
\hbox{J.~Gibson}~\cite{gibsonPerceptionVisualWorld1950} laid the foundation for studies of the way optical information guides animal movement.  Building on this, subsequent work in perceptual psychology studied \emph{time-to-contact} as a visual cue that plausibly guides animal movement as well as human motor vehicle operation 
\cite{leeTheoryVisualControl1976, leePlummetingGannetsParadigm1981,
wangTimeCollisionSignalled1992, srinivasanBiologyEngineeringInsect2012}.
Steering based on perceived optical flow has been explored in a relatively small set of robotics studies (reviewed in~\cite{raudies2012review}), including biologically-plausible estimators that replicate the visual navigation of flying insects~\cite{shoemakerOpticFlowEstimation2011}. With the rise of deep learning, dense optical-flow methods such as RAFT~\cite{teedRAFTRecurrentAllPairs2020}, FlowFormer++~\cite{shiFlowFormerMaskedCost2023}, and GMFlow~\cite{xuGMFlowLearningOptical2022} have become far more robust, but remain too computationally demanding for lightweight real-time navigation; the work below instead distills optical-flow sensing into the simple perceptual cue, TTT.

\myparagraph{TTT-based navigation}
Inspired by the three-dimensional movements of bats and other flying animals, steering control laws based on TTT were introduced in~\cite{zhaodankongOpticalFlowSensing2013}. 
Later,
~\cite{baillieulVisualNavigation2pixel2020} extended TTT usage to rotational motion by introducing the abstractions of Eulerian and Lagrangian optical flow.
To apply these concepts to robot navigation, previous work
~\cite{borettiVisualNavigationUsing2022} proposed a real-time optical-flow approach using the LK algorithm. To reduce noise, their method averages time-to-transit values across a few (i.e., three or five) fixed, environment-independent selected regions of interest, leaving parts of the visual field unused. They provide theoretical results on corridor-centering and single-wall-following, laying the foundation for navigation algorithms tested on several robotic platforms. However, since the LK optical-flow estimation is heavily affected by rotational motion, they introduced a \textit{sense-act} cycle that restricts estimation to periodic straight-line segments of the trajectory, significantly reducing the reactivity of the controller to more complex geometries.
\myparagraph{Other approaches to monocular camera depth perception}  Since~\cite{borettiVisualNavigationUsing2022}, monocular depth inference has been dominated by foundation models that recover scene geometry in fine detail, such as Depth-Anything-V2~\cite{yang2024depthanythingv2} and the recent Depth-Anything-3 (DA3)~\cite{lin2025depth}, while end-to-end navigation has shifted toward large Vision-Language-Action (VLA) models such as ViNT~\cite{shah2023vint}, NoMaD~\cite{sridhar2024nomad}, and OpenVLA~\cite{kim2024openvla}. These models deliver remarkable generality, but at the cost of billions of parameters and server-class GPUs. We deliberately move in the opposite direction: exploiting the fact that TTT is a \emph{coarse} cue, we obtain a $1.95$M-parameter estimator that runs on a Jetson Orin Nano, and we quantify the resulting scale gap in Sec.~\ref{sec:training}.

\subsection{Geometric and Perceived Time-to-Transit}
\label{sec:ttt}
To illustrate the notion of Time-to-Transit (TTT), we consider a monocular camera mounted on a robot that moves with planar unicycle kinematics:
\begin{equation}
\left(\begin{array}{c}
\dot x \\
\dot y \\
\dot\theta\end{array}\right) = \left(\begin{array}{c}
v\cos\theta \\
v\sin\theta \\
u\end{array}\right),
\label{eq:jb:BasicVehicle}
\end{equation}
where $v$ is the speed of the vehicle, $\theta$ is the angle between the \hbox{$x_r$-axis} of the robot body frame and the \hbox{$x$-axis} of a world frame, and $u$ is the turning rate.

\begin{figure}[]
    \centering
    \includegraphics[width=0.5\columnwidth]{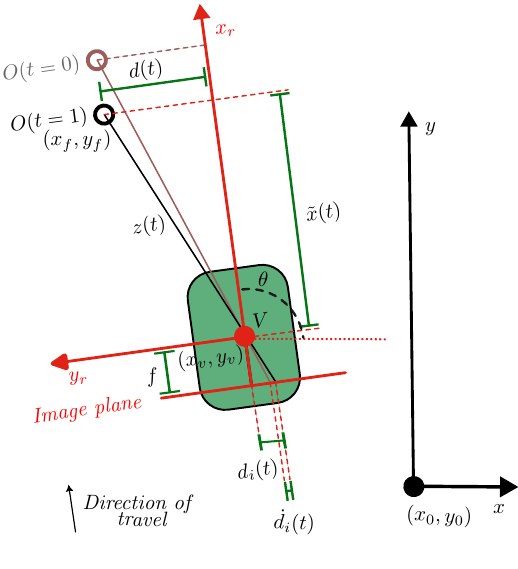}
    \vspace{-.6cm}
    \caption{Simple plane geometry representing a vehicle whose center is in $(x_v, y_v)$, moving at a constant speed $v$. The quantity $\theta$ represents the angle between the \hbox{$x_r$-axis} of the reference frame fixed to the robot and the \hbox{$x$-axis} of a global reference frame. The point $(x_f, y_f)$ indicates the position of a feature $O$ whose distance from the vehicle is described over time by $z(t)$. The time-to-transit value measured by the vehicle is $\tau_g=\frac{\tilde{x}}{v}$.}
    \label{fig:kinem2}
\end{figure}

The TTT visual cue has a simple geometric interpretation (Fig.~\ref{fig:kinem2}). Specifically, as described in \cite{baillieulPerceptualControlLarge2019, sebestaAnimalinspiredAgileFlight2012,zhaodankongOpticalFlowSensing2013,baillieulVisualNavigation2pixel2020}, if a vehicle is moving in a straight line and a feature point lies ahead of the camera, 
we can draw a plane that passes through the feature point and is also perpendicular to the line of travel.
The time-to-transit feature is defined as the time it will take the vehicle to follow the straight path at the current speed until it reaches this plane.
Following the definition, simple plane geometry (Fig.~\ref{fig:kinem2}) shows that the \emph{geometric} time-to-transit~\cite{baillieulPerceptualControlLarge2019} is given by:
\begin{equation}
    \tau_g(t) = \frac{\tilde{x}(t)}{v} = \frac{cos\theta(x_f-x_v(t))+sin\theta(y_f-y_v(t))}{v},
    \label{eq:taug}
\end{equation}
where $(x_f, y_f)$ is the position of a static feature in the global reference frame.
The geometric TTT value $\tau_g$ can be computed in retinotopic image coordinates 
under certain assumptions.

\begin{figure}
\centering
\includegraphics[width=0.8\columnwidth]{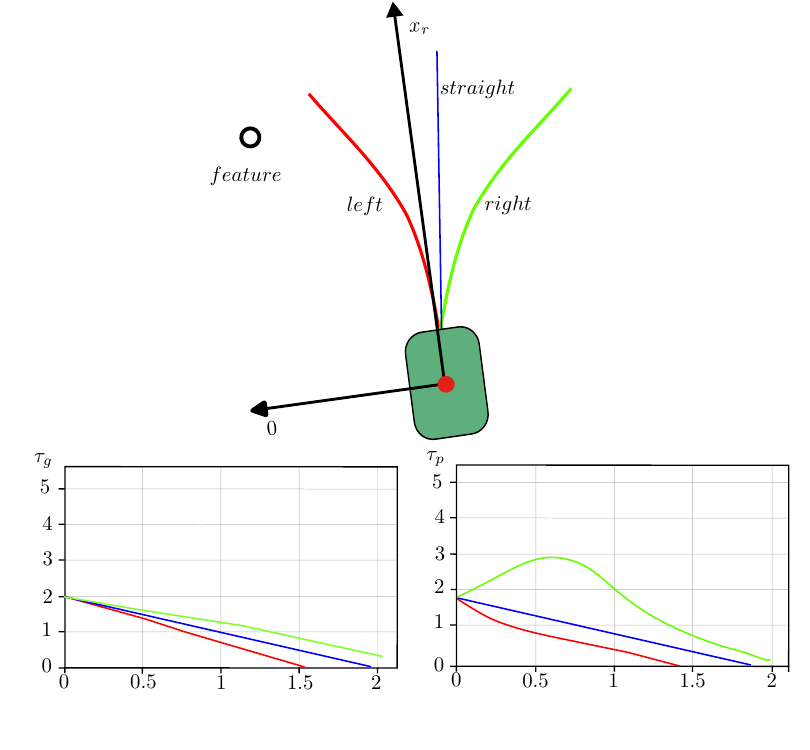}
\caption{Even a modest camera rotation can significantly distort the perceived time-to-transit. The unit-speed unicycle vehicle defined in~(\ref{eq:jb:BasicVehicle}) moves to transit a point feature that is to its left and ahead.  If the vehicle turns away from the feature, both $\tau_g(t)$ and $\tau_P(t)$ will decrease relative to following a straight path.  If the vehicle turns toward the feature, on the other hand, both values increase and $\tau_p(t)$ can be badly distorted.}
\label{fig:taugvstaup}
\end{figure}

Specifically, assuming a forward-looking camera, we define the \emph{lateral distance} $d(t) = D$ from the designated feature point. From similarity of triangles in Fig.~\ref{fig:kinem2}, it follows that:
\begin{equation}
    \frac{d(t)}{\tilde x(t)} = \frac{D}{\tilde x(t)} =  \frac{d_i(t)}{f}.
    \label{eq:base_for_ttt}
\end{equation}
From here, assuming furthermore a constant speed $v$, and constant heading $\theta(t)=\theta_0$, we have
\begin{equation}
    \label{eq:didot}
    \dot{d_i}(t)\tilde{x}(t)-d_i(t)v = 0;
\end{equation}
rearranging terms, we define the \emph{perceived} TTT as from the image point movement on the optical sensor:
\begin{equation}
    \label{eq:tau_per}
    \tau_p=\frac{\tilde{x}(t)}{v} = \frac{d_i(t)}{\dot{d_i}(t)}.
\end{equation}
While the \emph{geometric TTT} $\tau_g$ is required for navigation, it coincides with the \emph{perceived TTT} $\tau_p$, i.e., $\tau_g=\tau_p$, only under the assumptions above. See Fig.~\ref{fig:taugvstaup} for a graphical representation of the distortion introduced by rotations.
This is the main reason why~\cite{borettiVisualNavigationUsing2022} requires straight segments in a periodic \emph{sense-act} cycle. Unfortunately, with this strategy, estimation and rotation cannot be performed at the same time, forcing the control to be applied in open loop. 

\section{MobileNeTTT: Selection of a dnn-based TTT estimator}
\label{sec:dnn}
\begin{figure*}
\centering
\includegraphics[width=6in]{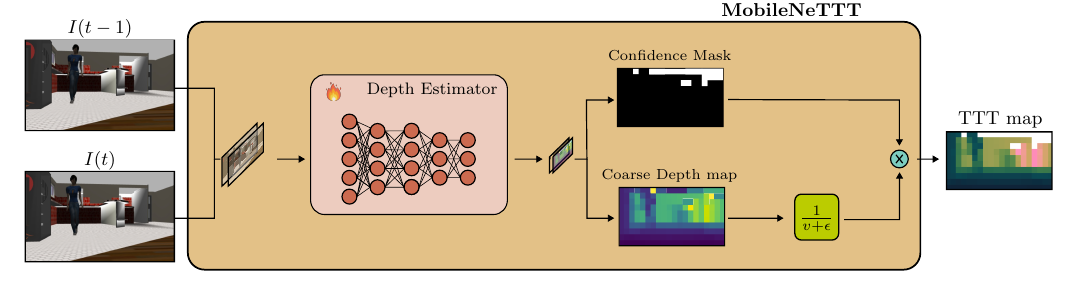}
\vspace{-.2cm}
\caption{Overview of MobileNeTTT, which estimates Time-To-Transit from sequential RGB images \( I(t-1) \) and \( I(t) \) using a DNN-based depth estimator. Instead of predicting a dense per-pixel depth map, the network outputs a coarse depth map and a binary validity map indicating reliable values. Coarse depth values are then divided by the robot's speed \( v \), with a small constant \( \epsilon \) added to prevent division by zero, to compute TTT values.
}
\label{fig:general_scheme}
\end{figure*}

In this section, we introduce MobileNeTTT, a time-to-transit estimator relying on DNN-based coarse depth estimation instead of sparse optical flow. 
Specifically, MobileNeTTT predicts the depth $\tilde{x}(t)$, and computes the geometric TTT using definition \eqref{eq:taug}, i.e., $\tau_g(t) = \tilde{x}(t)/v$.
MobileNeTTT balances robustness to noise and efficiency by estimating depth on a mesoscale grid partition with an associated confidence mask (Fig.~\ref{fig:general_scheme}).
The coarse partition maintains full visual field coverage without performing dense per-pixel depth prediction, but still allows pooling information across Regions of Interest (ROIs, as in~\cite{borettiVisualNavigationUsing2022}) that can also be defined dynamically (see Section~\ref{sec:dynamic_roi}). The mesoscale resolution is tuned to be suitable for real-time execution on edge hardware.
The confidence mask pairs each rectangular cell with a reliability value accounting for regions with infinite depth where no visible objects are present.\\
Once the required input and output have been set, we systematically evaluated multiple architectural variants and we define the final pipeline.

\myparagraph{Model Training} We train the depth estimator module in MobileNeTTT to produce a two-channel tensor containing the depth values $\tilde{x}(t)$ and the confidence mask. By focusing on this module, we remove dependency on the robot’s speed $v$, promote the learning of strong geometric features invariant to motion, and we can use existing large datasets for better training, performance, and generalization. Inspired by YOLO~\cite{tervenComprehensiveReviewYOLO2023}, we define a custom loss:

\begin{small}
\begin{equation}
\begin{aligned}
    \mathcal{L} &= \lambda_{TTT} \sum_{i=0}^{H} \sum_{j=0}^{W} C^{gt}_{ij} (\tilde{X}^{gt}_{ij} - \tilde{X}^{est}_{ij})^2 +\\
    &+ \lambda_{data} \sum_{i=0}^{H} \sum_{j=0}^{W} C^{gt}_{ij} (C^{gt}_{ij} - C^{est}_{ij})^2 +\\
    &+ \lambda_{nodata} \sum_{i=0}^{H} \sum_{j=0}^{W} (1 - C^{gt}_{ij}) (C^{gt}_{ij} - C^{est}_{ij})^2
\end{aligned}
\end{equation}
\end{small}

Here, $H=10$, $W=20$ represent the shape of the coarse depth map, $C^{gt}\in\{0,1\}^{H \times W}$ is the ground-truth confidence mask, and $\tilde{X}^{gt}\in\mathbb{R}^{H \times W}$ the corresponding $\tilde{x}$ map; $C^{est}$ and $\tilde{X}^{est}$ are the respective predictions. The weights $\lambda_{TTT}$, $\lambda_{data}$, and $\lambda_{nodata}$ balance the loss terms, with $\lambda_{nodata}$ typically larger to penalize false positives on regions without available depth. Focal loss is a viable alternative to further adjust the contribution of uncertain predictions.

\subsection{Datasets}
\label{sec:datasets}
Since we train the model to predict depth rather than TTT directly, we can leverage existing datasets. We use KITTI~\cite{geigerVisionMeetsRobotics2013} and RGBD-SLAM~\cite{sturmBenchmarkEvaluationRGBD2012}, which provide real-world urban and indoor scenes, respectively. We adapted these datasets to associate consecutive image pairs with a coarse $H \times W$ depth map and a binary confidence mask: each depth cell contains the average normalized depth (ranging from 0 to 1 according to the LiDAR range), while the mask indicates valid measurements ($1$) or missing data ($0$).
To fairly compare our TTT estimator against the OF-based method in~\cite{borettiVisualNavigationUsing2022}, we need consecutive image pairs without rotational motion, adhering to the sense-act cycle introduced in~\cite{borettiVisualNavigationUsing2022}. This assumption is generally not met in standard datasets, hence we generated a custom synthetic dataset recorded in Gazebo~\cite{koenigDesignUseParadigms2004a}.

\myparagraph{KITTI~\cite{geigerVisionMeetsRobotics2013}} A widely used autonomous-driving dataset; we focus on the \textit{Residential} section (roads with pedestrians, bicycles, and cars), using 18,248 images split into 14,320 for training/validation (80\%/20\%) and 3,928 for testing.

\myparagraph{RGBD-SLAM~\cite{sturmBenchmarkEvaluationRGBD2012}} Indoor videos with Kinect depth as ground truth; we use 11 sequences from the \textit{Handheld} and \textit{Robot SLAM} categories (excluding 4 with a fixed rotating or floor-pointing camera): 10,803 training, 2,700 validation, and 3,456 testing images.

\myparagraph{Custom Dataset} 
 The dataset we created specifically to train MobileNeTTT includes 17,612 images at $640 \times 480$ resolution. Of these, 14,612 are used for training (80\%) and validation (20\%), while 3,000 are reserved for testing in unseen environments.
Data were collected in Gazebo~\cite{koenigDesignUseParadigms2004a} simulated environments using a monocular RGB camera and a 3D LiDAR mounted at different heights on a Jackal robot. Some environments were custom-built, while others were sourced from external repositories\footnote{External environments found here: \href{https://github.com/leonhartyao/gazebo_models_worlds_collection}{https://github.com/gazebomodels}
.}. LiDAR point clouds provide ground-truth depth, projected onto the image plane via calibrated 3D-to-2D alignment, using the method from~\cite{koideGeneralSingleshotTargetless2023}. Some recordings are collected satisfying the assumptions required for the OF-based TTT estimator, and we use them to enable a fair comparison between the techniques. The dataset is publicly released\footnote{Dataset released upon paper acceptance.}.

\subsection{Metrics}
Our goal is to select an architecture by analyzing not only estimation accuracy, but also memory footprint (measured by the number of parameters M and memory usage in MB) and inference time (measured in milliseconds).
We evaluate the performance of the different depth estimators on two tasks: coarse depth estimation and binary confidence mask prediction.
For coarse depth estimation, we adopt standard metrics from \cite{zhaoMonocularDepthEstimation2020}: \textbf{Root Mean Square Error (RMSE)} and \textbf{Absolute Relative Error (ARE)} assess the overall prediction accuracy, while \textbf{Threshold Accuracy} measures the fraction of predictions within error bounds $\epsilon=1.25$, $1.25^2$, and $1.25^3$, denoted as $\delta_1$, $\delta_2$, and $\delta_3$.
To evaluate the confidence mask prediction, where class imbalance is prominent due to a higher prevalence of 1s (valid LiDAR data), we use the \textbf{Macro-averaged F1 score} \cite{sokolovaSystematicAnalysisPerformance2009}. This metric averages the F1 scores of both classes equally, offering a balanced view by accounting for both precision and recall, regardless of class frequency.

\subsection{Architecture comparison and selection}
\label{sec:training}
The simplified nature of coarse depth estimation, allows us to adapt several lightweight DNN architectures, even those not originally designed for depth tasks.
Given our goal of building a lightweight framework for real-time applications, we trained and compared several estimator modules: \textit{\textbf{MobileNet-v3 Large}}~\cite{howardSearchingMobileNetV32019}, \textit{\textbf{ResNet-50}}~\cite{heDeepResidualLearning2016}, \textit{\textbf{MobileViT-xs}}~\cite{mehtaMobileViTLightweightGeneralpurpose2022}, and \textit{\textbf{GMDepth}}~\cite{xuUnifyingFlowStereo2023}.
Our model, MobileNeTTT, uses the \textit{\textbf{MobileNet-v3 Small}} architecture~\cite{howardSearchingMobileNetV32019} augmented with a channel attention block along the lines of \cite{wooCBAMConvolutionalBlock2018}.

In this section, we compare the performance of these different coarse depth estimators on the datasets and on the metrics described above. 
\begin{table*}[]
\centering
\resizebox{0.9\textwidth}{!}{
 \begin{tabular}{llcccccc}

    \toprule
                & & \begin{tabular}{c}RMSE $\downarrow$\end{tabular} & ARE $\downarrow$ & \begin{tabular}{c}$\delta_1 \uparrow$ \end{tabular} & \begin{tabular}{c}$\delta_2 \uparrow$ \end{tabular} & \begin{tabular}{c}$\delta_3 \uparrow$ \end{tabular} & \begin{tabular}{c}F1 Macro $\uparrow$\end{tabular}\\
    \midrule

    \multirow{5}{*}{\rotatebox{90}{\scriptsize{Synthetic}}}
    & ResNet-50                     & 0.129 \tiny{$\pm$ 0.003}             & 0.170 \tiny{$\pm$ 0.002}           & 76.2\% \tiny{$\pm$ 0.3\%}           & 88.8\% \tiny{$\pm$ 0.1\%}           & 93.9\% \tiny{$\pm$ 0.1\% }          & 84.1\% \tiny{$\pm$ 0.1\%}              \\
    & MobileViT-xs                  & 0.128 \tiny{$\pm$ 0.004}             & 0.165 \tiny{$\pm$ 0.005}           & 80.2\% \tiny{$\pm$ 0.1\%}           & 90.7\% \tiny{$\pm$ 0.2\%}           & 94.6\%  \tiny{$\pm$ 0.1\%}          & 85.9\%  \tiny{$\pm$ 0.1\%}            \\
    & MobileNet-Large               & \textbf{0.113 \tiny{$\pm$ 0.004}}    & 0.158 \tiny{$\pm$ 0.002}           & 79.9\% \tiny{$\pm$ 0.3\%}  & 90.5\% \tiny{$\pm$ 1.0\% }          & 94.5\% \tiny{$\pm$ 0.6\%}           & \textbf{86.9\% \tiny{$\pm$ 0.6\%}}      \\

     & GMDepth               & 0.144 \tiny{$\pm$ 0.005}             & 0.213 \tiny{$\pm$ 0.014}            & 71.1\% \tiny{$\pm$ 1.3\%}    & 85.9\% \tiny{$\pm$ 1.1\%}            & 93.4\% \tiny{$\pm$ 0.3\%}           & 76.0\% \tiny{$\pm$ 1.6\%}      \\
    & \cellcolor{customgray!20}MobileNeTTT
    & \cellcolor{customgray!20}\textbf{0.114 \tiny{$\pm$ 0.004}}             & \cellcolor{customgray!20}\textbf{$^*$0.152 \tiny{$\pm$ 0.003}}   & \cellcolor{customgray!20}\textbf{$^*$80.8\% \tiny{$\pm$ 0.4\%}}           & \cellcolor{customgray!20}\textbf{$^*$92.1\% \tiny{$\pm$ 0.5\%}}  & \cellcolor{customgray!20}\textbf{$^*$95.8\% \tiny{$\pm$ 0.4\%}}  & \cellcolor{customgray!20}\textbf{86.4\% \tiny{$\pm$ 0.1\%}}               \\
    \midrule
    \multirow{5}{*}{\rotatebox{90}{\scriptsize{RGBD-SLAM}}}
    & ResNet-50                     & 0.044 \tiny{$\pm$ 0.001}             & 0.219 \tiny{$\pm$ 0.003}            & 73.7\% \tiny{$\pm$ 0.4\%}            & 89.2\% \tiny{$\pm$ 0.3\%}            & 93.8\% \tiny{$\pm$ 0.3\%}           & 83.0\% \tiny{$\pm$ 0.1\%}              \\
    & MobileViT-xs                  & 0.038 \tiny{$\pm$ 0.001}              & 0.189 \tiny{$\pm$ 0.002}            & \textbf{$^*$79.4\% \tiny{$\pm$ 0.2\%}}            & \textbf{92.3\% \tiny{$\pm$ 0.2\%}}            & \textbf{96.2\% \tiny{$\pm$ 0.5\%}}           &83.9\% \tiny{$\pm$ 0.7\%}             \\
    & MobileNet-Large               & 0.038 \tiny{$\pm$ 0.001}             & 0.190 \tiny{$\pm$ 0.005}            & 77.4\% \tiny{$\pm$ 0.9\%}    & 91.3\% \tiny{$\pm$ 0.3\%}            & 94.9\% \tiny{$\pm$ 0.3\%}           & 84.5\% \tiny{$\pm$ 0.1\%}      \\
    & GMDepth               & 0.065 \tiny{$\pm$ 0.003}             & 0.379 \tiny{$\pm$ 0.008}            & 54.8\% \tiny{$\pm$ 0.7\%}    & 74.9\% \tiny{$\pm$ 1.2\%}            & 85.5\% \tiny{$\pm$ 1.7\%}           & 78.8\% \tiny{$\pm$ 1.2\%}      \\
    & \cellcolor{customgray!20}MobileNeTTT          &
    \cellcolor{customgray!20}\textbf{$^*$0.035 \tiny{$\pm$ 0.001}}     & \cellcolor{customgray!20}\textbf{$^*$0.185 \tiny{$\pm$ 0.002}}   & \cellcolor{customgray!20}76.9\% \tiny{$\pm$ 0.6\%}             & \cellcolor{customgray!20}\textbf{91.7\% \tiny{$\pm$ 0.6\%}}    &\cellcolor{customgray!20}\textbf{95.6\% \tiny{$\pm$ 0.2\%}}   & \cellcolor{customgray!20}\textbf{$^*$84.8\% \tiny{$\pm$ 0.1\%}}               \\
    \midrule
    \multirow{5}{*}{\rotatebox{90}{\scriptsize{KITTI}}}
    & ResNet-50                     & 0.333 \tiny{$\pm$ 0.010}             & 0.113 \tiny{$\pm$ 0.003}           & 89.2\% \tiny{$\pm$ 0.3\%}           & 94.7\% \tiny{$\pm$ 0.2\%}           & 95.8\% \tiny{$\pm$ 0.3\%}          & 95.0\% \tiny{$\pm$ 0.1\%}             \\
    & MobileViT-xs                  & 0.299 \tiny{$\pm$ 0.003}             & 0.114 \tiny{$\pm$ 0.001}           & \textbf{90.1\% \tiny{$\pm$ 0.2\%}}           & 94.5\% \tiny{$\pm$ 0.1\%}           & 95.6\% \tiny{$\pm$ 0.1\%}          & 94.5\%  \tiny{$\pm$ 0.2\%}            \\
    & MobileNet-Large               & 0.316 \tiny{$\pm$ 0.008}             & 0.114 \tiny{$\pm$ 0.003}           & \textbf{90.6\% \tiny{$\pm$ 0.5\%}}   & 94.7\% \tiny{$\pm$ 0.2\%}            & 96.2\% \tiny{$\pm$ 0.2\%}          & 95.1\% \tiny{$\pm$ 0.1\%}     \\
    & GMDepth               & 0.355 \tiny{$\pm$ 0.011}             & 0.142 \tiny{$\pm$ 0.002}            & 84.2\% \tiny{$\pm$ 0.3\%}    & 92.6\% \tiny{$\pm$ 0.5\%}            & 94.9\% \tiny{$\pm$ 0.4\%}           & 93.8\% \tiny{$\pm$ 0.2\%}      \\
    & \cellcolor{customgray!20}MobileNeTTT
    & \cellcolor{customgray!20}\textbf{$^*$0.292 \tiny{$\pm$ 0.002}}    & \cellcolor{customgray!20}\textbf{$^*$0.103 \tiny{$\pm$ 0.003}}  & \cellcolor{customgray!20}\textbf{90.2\% \tiny{$\pm$ 0.3\%}}            & \cellcolor{customgray!20}\textbf{$^*$95.2\% \tiny{$\pm$ 0.1\%}}   &\cellcolor{customgray!20}\textbf{$^*$96.7\% \tiny{$\pm$ 0.2\%}}  & \cellcolor{customgray!20}\textbf{$^*$95.8\% \tiny{$\pm$ 0.1\%}}              \\
    \bottomrule

  \end{tabular}
}
  \caption{Results on our synthetic dataset, RGBD-SLAM, and KITTI. Best results are in bold (considering overlapping standard deviation); unique bests are marked with $^*$. In light gray, we highlight the module selected for use in MobileNeTTT.
}
  \label{tab:simulation_results}
  \end{table*}

\myparagraph{Depth estimation performance}
Our custom synthetic dataset, collected using Gazebo simulations, was used to train all the tested neural network architectures. 
Starting from available pretrained models\footnote{Pretrained models were obtained from Hugging Face. For GMDepth, we use the code available at \hbox{\href{https://github.com/autonomousvision/unimatch}{https://github.com/unimatch}}.}, all networks were trained for 100 epochs with an initial learning rate of 0.001 using the Adam optimizer \cite{kingmaAdamMethodStochastic2017}.
Table~\ref{tab:simulation_results} shows the mean accuracy and standard deviation on the test set for the different depth estimators. Larger models like ResNet-50 or GMDepth performed worse on the coarse depth task, likely due to their complexity and requiring larger amounts of data to train effectively. More fundamentally, this is the central observation of this work, and it extends to the much larger foundation models of Sec.~\ref{sec:related_works}: the capacity such models devote to border effects, depth discontinuities, and occlusions aids per-pixel accuracy but is largely irrelevant, even a source of data hunger and overfitting, for the coarse full-field signal that TTT requires. Indeed, smaller models performed better overall, with our MobileNeTTT slightly ahead, achieving the best scores in ARE, $\delta_1$, $\delta_2$, and $\delta_3$, and matching MobileNet-Large in RMSE and F1 macro despite its small size.
The top model from each architecture, trained on our custom dataset, was fine-tuned on KITTI and RGBD-SLAM for 50 epochs using the Adam optimizer with a learning rate of 0.001. As shown in Table~\ref{tab:simulation_results}, MobileNeTTT consistently outperformed larger models, achieving the best RMSE and ARE, and matching or exceeding others in $\delta_1$, $\delta_2$, and $\delta_3$.

\begin{table}[]
\centering
 \begin{tabular}{lccc}
    \toprule
     Depth Estimator     & Param. (M)      & Memory (MB)          & Inf. Time (ms) \\
    \midrule
    ResNet-50            &  26.62          &  114.56               &   30.11        \\
    MobileNet-Large      &  5.22           &  20.97                &   14.72        \\
    GMDepth              &  4.68           &  18.72                &   112.96        \\
    MobileViT-xs         &  2.48           &  10.03                &   23.12        \\
    \cellcolor{customgray!20}\textbf{MobileNeTTT}          &  \cellcolor{customgray!20}\textbf{1.95}  &  \cellcolor{customgray!20}\textbf{7.86}        & \cellcolor{customgray!20}\textbf{10.27} \\
    \bottomrule
  \end{tabular}
\caption{Parameters, memory occupation and inference time (on Nvidia Jetson Orin Nano) of the coarse depth estimators.}

\label{tab:param}
\end{table}

\myparagraph{Number of parameters and inference time}
Given the importance of having a neural model that can run in real-time on resource-constrained robotic platforms, Table~\ref{tab:param} compares various models on the NVIDIA Jetson Orin Nano in terms of number of parameters, memory usage, and inference time. The results show MobileNeTTT as the best choice, achieving 97 Frames Per Second (FPS) while using less than half the memory of \hbox{MobileNet-Large} and running 2.2 times faster than \hbox{MobileViT-xs}. With comparable or better estimation quality than larger architectures, this model will be used in the following sections to evaluate improvements in time-to-transit estimation and navigation experiments.

\myparagraph{On-device implementation} The coarse $10\times20$ output is what makes real-time edge inference practical, and we pair it with a deployment pipeline that preserves the network predictions bit-for-bit. Each frame pair is written into a single page-locked buffer and transferred once to the GPU, where normalization is fused and the network runs in half precision. We then capture the entire forward pass as a CUDA graph, which removes the per-kernel launch overhead that dominates latency at batch size one. We further observed that the channel-attention block computed its global average with a full-image pooling kernel, and replaced it with an equivalent direct spatial mean. All of these optimizations are accuracy-preserving: the optimized model reproduces the coarse depth and confidence-mask outputs bit-for-bit.

\myparagraph{Relation to large foundation models}
We do not aim to match the general scene-understanding or language-grounded capabilities of large depth and navigation models; we target a single coarse cue, a full-field TTT estimate computed in real time on edge hardware, and show a $1.95$M-parameter network is sufficient. The resulting resource gap is decisive and hardware-independent: at $1.95$M parameters and under $8$\,MB, MobileNeTTT is roughly $170\times$ smaller than Depth-Anything-V2-Large ($335$M, $1.3$\,GB) and $560$ to $4000\times$ smaller than DA3 ($\sim$$1.1$B, $\sim$$4.4$\,GB) and $7$ to $8$B-parameter VLAs such as OpenVLA ($\sim$$30$\,GB), which require server-class accelerators that reactive TTT navigation does not.

\subsection{Comparison with Optical Flow for Time-To-Transit estimation quality}
\label{sec:ttt_est_qual}

\begin{figure}
\centering
\includegraphics[width=0.95\columnwidth]{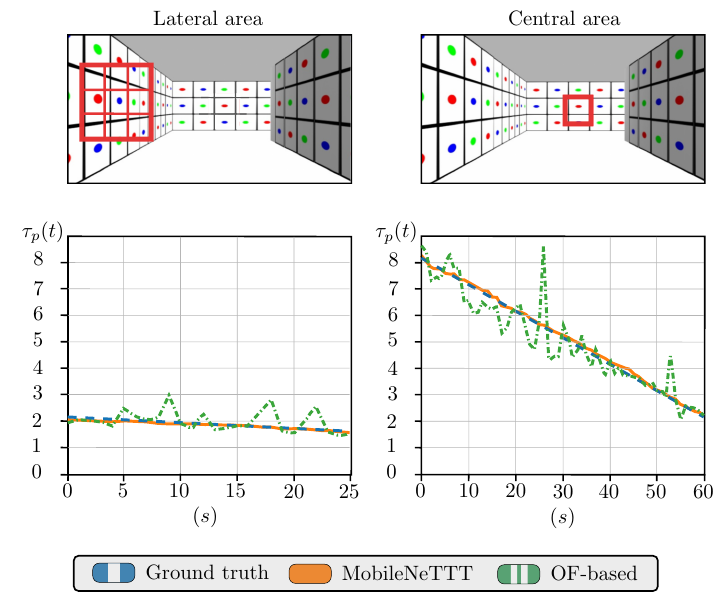}
\vspace{-.2cm}
\caption{Comparison of the estimation of TTT values coming from two areas of the input image using MobileNeTTT and the OF-based method. The image is acquired in a simulated environment. The red outer area represents the Region of Interest (ROI) where features are selected and tracked using optical flow. These ROIs are combinations of the grid cells provided by MobileNeTTT. MobileNeTTT offers superior, less noisy estimates, particularly in the central region close to the FoE.}
\label{fig:ttt_nn_of}
\end{figure}

Having trained MobileNeTTT as a depth estimator, we convert it to perform TTT estimation by dividing depth by speed as illustrated in
Figure~\ref{fig:general_scheme}, and compare it to the Optical Flow-based approach from~\cite{borettiVisualNavigationUsing2022}, which estimates TTT as $\tau = \frac{d_i(t)}{\dot{d_i}(t)}$. The key difference lies in the estimation strategy: the Optical Flow method relies on tracking independent features to compute $d_i(t)$ and $\dot{d_i}(t)$, making it highly sensitive to occlusions and ineffective in regions lacking trackable features. In contrast, MobileNeTTT, leveraging the depth estimator can infer missing information from neighboring grid cells, enabling robust estimation even in sparse or occluded areas.
Additionally, the Optical Flow method performs poorly near the Focus of Expansion (FoE), where small errors in flow estimation lead to significant TTT inaccuracies.
To illustrate this, we estimate TTT using both methods and then aggregate the estimated values in the Regions-of-Interest (ROIs)
from~\cite{borettiVisualNavigationUsing2022}. 
Figure~\ref{fig:ttt_nn_of} shows the results in a corridor environment, captured by a monocular camera mounted on a Jackal robot moving without rotation\footnote{This environment was not previously seen by MobileNeTTT.}.
In particular we show the estimation of TTT in the lateral region of the input image (Figure~\ref{fig:ttt_nn_of}, left) and in the central region (Figure~\ref{fig:ttt_nn_of}, right).
As expected, the Optical Flow method is significantly noisier, particularly near the Focus of Expansion (FoE). 

Providing a quantitative comparison using the standard  datasets  described in Section~\ref{sec:datasets} is not straightforward, as correctly applying the Optical Flow-based estimation method that uses the \emph{sense-act} cycle requires that consecutive image pairs in the dataset are acquired in the absence of robot rotations. To ensure this condition is met, we instead rely on the test set of our synthetic dataset presented in Section ~\ref{sec:datasets}, which allows us to precisely identify scenarios where such an assumption holds. In Table~\ref{tab:rotation_comparison}, we report the complete results both on the entire dataset and on the subset of scenarios where the operational assumptions (i.e.\ no rotation) of the Optical Flow-based method are satisfied.
\begin{table}[t]
\centering
\begin{tabular}{ll|cc}
\toprule
\textbf{Scenario} & \textbf{Metric} & \textbf{OF-based}~\cite{borettiVisualNavigationUsing2022} & \textbf{MobileNeTTT} \\
\midrule
\multirow{2}{*}{w/o rotations}
  & MSE  $\downarrow$   & $1.40 \pm \text{\tiny 0.75}$         & $\mathbf{0.01} \pm \text{\tiny $\leq$0.01}$ \\
  & Wasserstein $\downarrow$  & $1.59 \pm \text{\tiny 0.72}$         & $\mathbf{0.10} \pm \text{\tiny 0.07}$ \\
\midrule
\multirow{2}{*}{with rotations}
  & MSE $\downarrow$      & $7.42 \pm \text{\tiny 2.18}$         & $\mathbf{0.02} \pm \text{\tiny $\leq$0.01}$ \\
  & Wasserstein $\downarrow$ & $2.28 \pm \text{\tiny 1.12}$         & $\mathbf{0.31} \pm \text{\tiny 0.11}$ \\
\bottomrule
\end{tabular}
\caption{Comparison of TTT estimation quality between the OF-based method and MobileNeTTT on our synthetic dataset test set.
}
\label{tab:rotation_comparison}
\end{table}
To evaluate the effectiveness of the considered methods, we adopt the same reference grid defined by MobileNeTTT and introduce two evaluation metrics: the mean squared error (MSE) and the Wasserstein distance among mean TTT values along the horizontal axis, as TTT-based control primarily focuses on lateral balancing. In both cases, MobileNeTTT achieves significantly better performance.

\section{Vision-based steering using MobileNeTTT}
\label{sec:vision_steering}
We present here a simple example to show how these improvements can also positively impact navigation. With a better understanding of the environment, it becomes possible to switch more accurately between control laws and enhance the performance of existing control strategies. In the next subsections, we briefly recall the TTT-based control laws, as proposed in~\cite{borettiVisualNavigationUsing2022}, and present an example of how better estimation can lead to improved navigation performance.

\subsection{Tau Balancing and Single Wall strategy}
A camera-equipped mobile robot with the kinematics described in (\ref{eq:jb:BasicVehicle}) moving at a certain speed \(v\) can be guided in a corridor with different steering laws using time-to-transit average values, \(\tau_r\) and \(\tau_l\), coming from fixed areas from the right and the left part of the input image, respectively.
In~\cite{baillieulPerceptualControlLarge2019}, the author presents \textit{Tau Balancing} for the vehicle kinematics~(\ref{eq:jb:BasicVehicle}) as
\begin{equation}
u(t) = k(\tau_l(t) - \tau_r(t))
\label{eq:tau_balance_eq}
\end{equation}
This centers the vehicle between walls in a corridor environment. If only values from one of the two regions are available, the robot can use the \textit{Single Wall} strategy from~\cite{borettiVisualNavigationUsing2022} defined as
\begin{equation}
u(t) = \pm k(\tau_x(t) - c)
\label{eq:tau_balance_const_eq}
\end{equation}
where $k$ is a positive gain with the sign determined by which side of the image $\tau$ is coming from and $c$ is a parameter that determines the steady state distance to the wall on which the observed features are located.

\subsection{Environment-Responsive Dynamic Regions of Interest}
\label{sec:dynamic_roi}
\begin{figure}
\centering
\includegraphics[width=0.85\columnwidth]{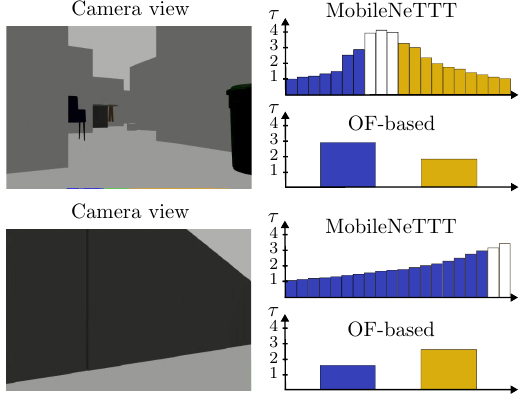}
\caption{An example of an environment and the TTT estimates produced by the OF-based method and MobileNeTTT. MobileNeTTT enables a better understanding of the environment and the definition of dynamic left (blue), right (orange), and central (white) ROIs. 
}
\label{fig:hist_ttt}
\end{figure}

Due to the estimation noise of TTT, the OF-based method in \cite{borettiVisualNavigationUsing2022} produces only two TTT values, which are the averages of estimates from two fixed areas. This limits the understanding of the environment's geometry, leading to two interconnected issues for navigation: 1) incorrect switching between control laws and 2) reliance on fixed ROIs for control. In Fig.~\ref{fig:hist_ttt}, we show an example of the input image, with, on the right, the output of the OF-based method (bottom graph) and the vertically averaged output of MobileNeTTT (top graph). Thanks to improved estimation, it is possible to generate an output that offers more detailed information about the surrounding environment. Furthermore, existing control laws can still be applied, while leveraging TTT values to create dynamic perceptions of affordances for turning to avoid obstacles in the environment.
In the experiments described below, as well as in Fig.~\ref{fig:hist_ttt}, the left, center, and right areas are defined based on the TTT distribution. First, the column with the maximum TTT value is identified. Starting from this peak, we search both to the left and to the right for the nearest columns whose TTT estimate is at least 10\% lower than the maximum value. These two columns define the boundaries of the center area, while all columns outside these boundaries are assigned to the left and right areas, respectively.

\subsection{Navigation Improvements}

\begin{figure}
\centering
\includegraphics[width=0.82\columnwidth]{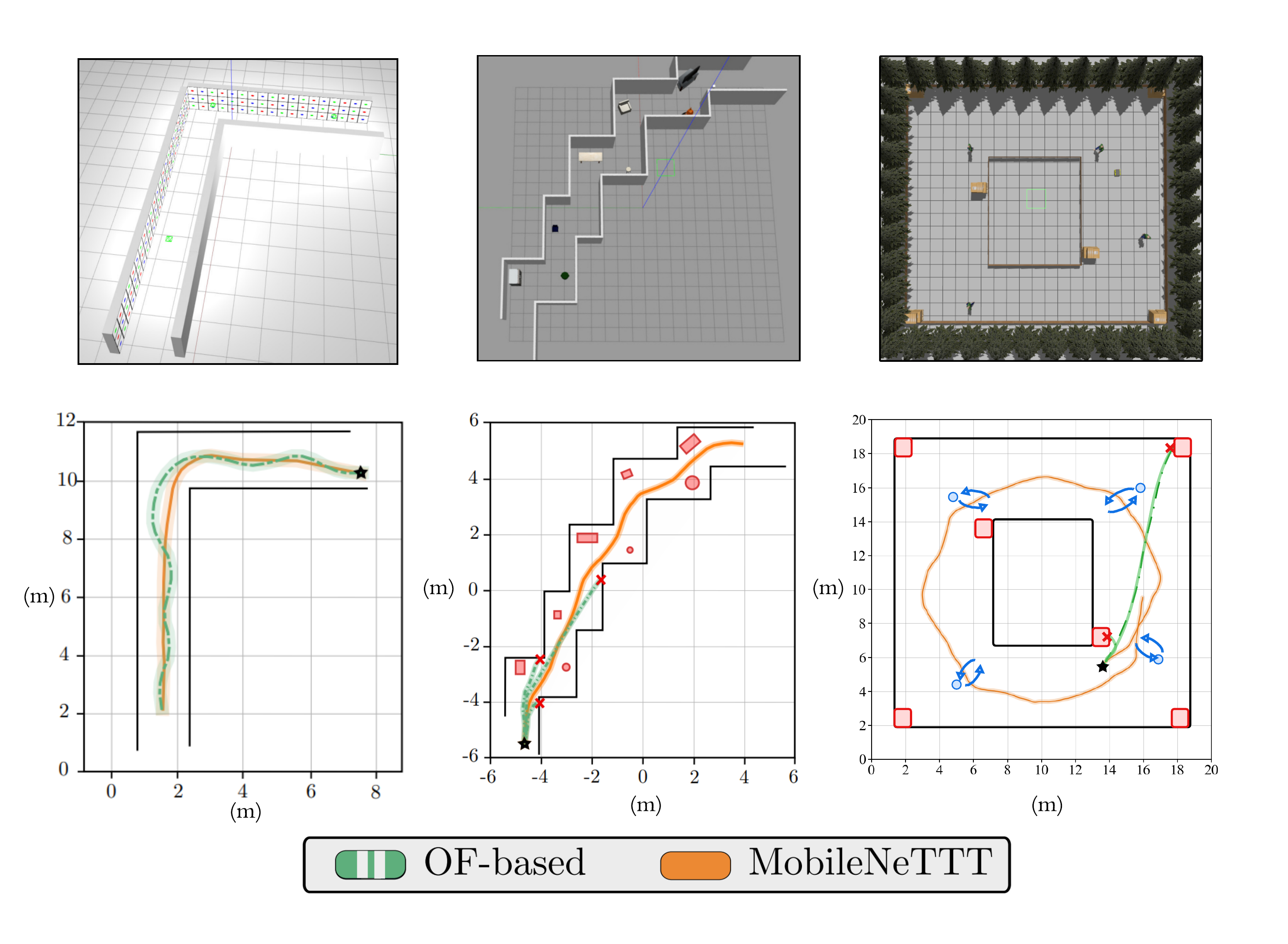}
\caption{Navigation trajectories produced by the OF-based method (green)
and MobileNeTTT (orange) across three experimental environments:
(left) an L-shaped corridor, (center) a winding corridor with multiple
successive alternating turns, and (right) an obstacle course environment.
The bottom panels show the corresponding odometry trajectories,
demonstrating MobileNeTTT's ability to navigate more smoothly and avoid
collisions compared to the OF-based method.
}
\label{fig:simulation_results}
\end{figure}

Using the controller of \cite{baillieulPerceptualControlLarge2019} with MobileNeTTT and dynamic ROIs, we compare against the OF-based method across several scenarios (Fig.~\ref{fig:simulation_results}), including an emergent capability of navigating around obstacles. In an L-shaped corridor (left), OF-based estimation causes oscillations because the \emph{sense-act} sensing phase must run without rotating the robot, whereas MobileNeTTT yields a smoother trajectory that stays near the corridor center.

In a zig-zag corridor with obstacles and featureless walls (Fig.~\ref{fig:simulation_results}, center), the OF-based method~\cite{borettiVisualNavigationUsing2022} receives no information from the untextured walls and collides, with behavior sensitive to the starting position; MobileNeTTT instead infers TTT from feature-rich neighboring cells and reliably reaches the end of the corridor.
In a dynamic environment with walking agents (Fig.~\ref{fig:simulation_results}, right), MobileNeTTT detects affordances around both static objects and moving agents and navigates without collision across repeated trials, unlike the OF-based method~\cite{borettiVisualNavigationUsing2022}.
Across all scenarios, Table~\ref{tab:nav} quantifies these trends: under identical control laws and gains, with randomized start positions over $N$ trials per environment, we report the goal-reaching success rate and the number of collisions for the OF-based controller and for MobileNeTTT. The OF-based method results in a high number of collisions across all tested scenarios, whereas MobileNeTTT achieves a high success rate, ranging from 80\% to 100\%, in both simulated and real-world environments. Importantly, the neural network was trained exclusively on simulated scenarios and evaluated in real-world environments that differ substantially from those encountered during training or fine-tuning.  Ongoing work is adding real-world experimental data to the training and test sets.

\begin{table}[t]
\centering
\begin{tabular}{lcccc}
\toprule
\multirow{2}{*}{\textbf{Environment} ($N=10$ trials)} & \multicolumn{2}{c}{\textbf{OF-based}~\cite{borettiVisualNavigationUsing2022}} & \multicolumn{2}{c}{\textbf{MobileNeTTT}} \\
\cmidrule(lr){2-3}\cmidrule(lr){4-5}
 & Succ.\% & Coll. & Succ.\% & Coll. \\
\midrule
L-shaped corridor          & 50\% & 5/10 & 100\% & 0/10 \\
Obstacle course            & 0\% & 10/10 & 100\% & 0/10 \\
Dynamic (walking agents)   & 0\% & 10/10 & 100\% & 0/10 \\
Real-World Obstacle Course \, $(N=5)$ & 20\% & 4/5 & 80\% & 1/5 \\
\bottomrule
\end{tabular}
\caption{
Goal-reaching success rates and collision counts for the OF-based
baseline and MobileNeTTT across the selected environments.
Both methods used the same controller parameters,
with steering gain range $k=0.6 - 0.8$ and forward speed $v=0.8 \frac{m}{s}$.  For the real laboratory experiment, comparisons were done with a gain of $k = 0.5$ and $v=0.3 \frac{m}{s}$.
}
\label{tab:nav}
\end{table}

\subsection{Robustness Testing and Experimental Results}
\begin{figure}
\centering
\includegraphics[width=0.82\columnwidth]{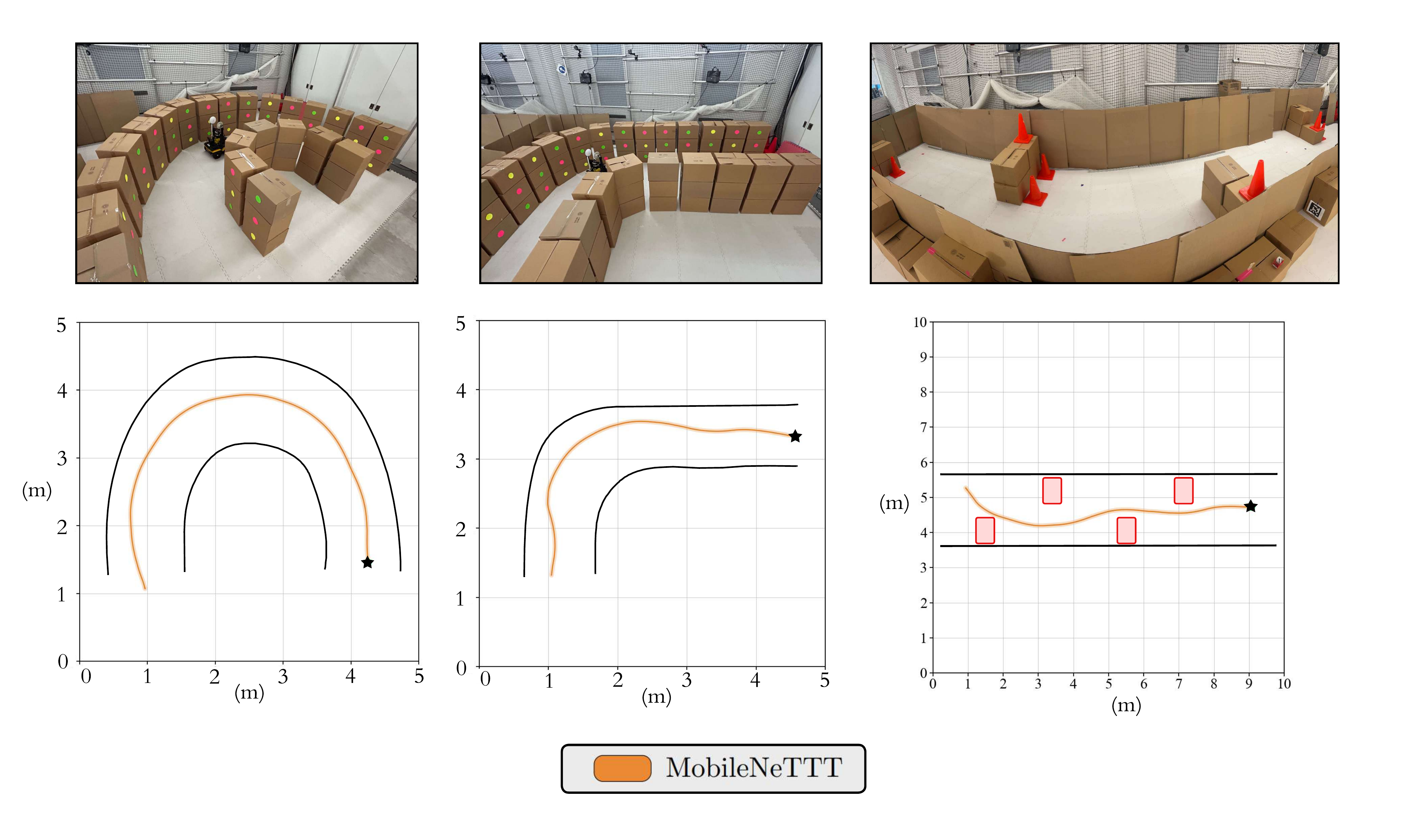}
\caption{Physical experiments conducted in an indoor robotics lab.
(Top) Three views of the real-world corridor constructed using cardboard
boxes, including a winding corridor with obstacles. (Bottom) The corresponding
top-down environment layout with the recorded MobileNeTTT odometry trajectory,
demonstrating successful navigation through the corridor without collision at
$v = 0.4\frac{m}{s}$.}

\label{fig:experiment_results}
\end{figure}

In addition to the experiments in simulated environments, MobileNeTTT and our dynamic ROI algorithm have also been tested on a Clearpath Jackal robot moving in real-case scenarios within an indoor robotics laboratory environment
(seen in the top panels of Fig. \ref{fig:experiment_results}).
The Jackal carries an Intel i7-9700 CPU, a GeForce GTX 1650 Ti GPU, and an Intel RealSense D435 camera (used for monocular RGB images).
The experiments involve steering laws presented in Eqs. \ref{eq:tau_balance_eq} and
\ref{eq:tau_balance_const_eq}
with stationary obstacles placed at random positions
protruding from the corridor walls as depicted in the right-hand panel of Fig. \ref{fig:experiment_results}.
For a constant vehicle speed of $0.4\frac{m}{s}$ and corridor width $1.5$ meters, effective gains for the steering law in Eq. \ref{eq:tau_balance_eq} were found to be in the range $k=0.3$ to $k=0.5$.

As demonstrated from odometry data reported in graphs in the lower panels of Fig. \ref{fig:experiment_results}, the control laws based on dynamic ROIs and using the TTT estimated with MobileNeTTT steer the robot toward the center of the corridor while also avoiding obstacles.

\section{Conclusion}
\label{sec:conclusion}
In this letter we introduced MobileNeTTT, a lightweight DNN-based framework for real-time time-to-transit estimation in robotic navigation. By casting TTT as a coarse depth-estimation problem, our approach eliminates optical flow computation while, unlike the prior LK-based method, providing full-field estimates without a sense-act cycle and remaining robust to rotations, dynamic objects, and sparse visual regions.
We also showed how this enhanced perception benefits navigation, with improved TTT estimates enabling dynamic Regions-of-Interest (ROIs) for more stable control laws and for steering safely around obstacles. MobileNeTTT does not attempt the general capabilities of large VLAs; rather, with only $1.95$M parameters and under $8$\,MB it runs in real time on a Jetson Orin Nano, showing that reactive, bio-inspired navigation needs a compact coarse cue rather than a data-center-scale model.

\myparagraph{Future work}
Operating in dynamic, sparse environments while producing broader TTT grids opens the door to more advanced control laws, potentially optimized with reinforcement learning.


\bibliography{./references.bib}
%
%
\bibliographystyle{IEEEtran}

\addtolength{\textheight}{-12cm}   

\end{document}